\documentclass[preprint,12pt]{elsarticle}

\usepackage{graphicx}
\usepackage{amssymb}
\usepackage[]{graphicx}
\usepackage{subfigure}
\usepackage[utf8]{inputenc}
\usepackage{setspace}
\usepackage[margin=1.25in]{geometry}
\graphicspath{ {figures/} }
\usepackage{subcaption}
\usepackage{amsmath}
\usepackage{multirow}
\usepackage{array}
\usepackage{booktabs}
\usepackage{algorithm}
\usepackage{algorithmic}
\usepackage{lineno}
\usepackage{fancyhdr}
\usepackage{lastpage}

\usepackage[numbers,sort&compress]{natbib}
\makeatletter
\let\ps@pprintTitle\ps@plain
\let\ps@pprintTitleCopy\ps@plain
\let\ps@frontmatter\ps@plain
\makeatother
\begin{document}
\begin{frontmatter}
\linenumbers
\title{Variable-frame CNNLSTM for Breast Nodule Classification using Ultrasound Videos}

\author[1]{Xiangxiang Cui\corref{cor1}}
\author[2]{Zhongyu Li\corref{cor1}}
\cortext[cor1]{Indicates equal contribution;}
\author[3]{Xiayue Fan}
\author[4]{Peng Huang}
\author[6]{Ying Wang}
\author[5]{Meng Yang}
\author[4]{Shi Chang}
\author[2]{Jihua Zhu}
\cortext[cor2]{Corresponding authors: P. Huang (xiangyahp@csu.edu.cn), J. Zhu (zhujh@xjtu.edu.cn)}

\address[1]{The State Key Lab of Cognitive Neuroscience and Learning, Beijing Normal University}
\address[2]{School of Software Engineering, Xi'an Jiaotong University}
\address[3]{Center for Immunological and Metabolic Diseases, The First Affiliated Hospital of Xi'an Jiaotong University}
\address[4]{Department of General Surgery, Xiangya Hospital, Central South University}
\address[5]{Frontline Intelligent Technology (Nanjing) Co., Ltd.}
\address[6]{Department of general surgery, The Second Hospital of Hebei Medical University}



\begin{abstract}

\textbf{Purpose:}
The intersection of medical imaging and artificial intelligence has become an important research direction in intelligent medical treatment, and deep learning can analyze and calculate medical images and participate in clinical diagnosis. In order to improve the dataset utilization, the method supporting the training of variable frame datasets and the automatic extraction of ultrasonic features combined with deep learning is designed to improve the dataset utilization rate, which has important research value in ultrasound classification diagnosis. There is no extraction of time series features for the existing keyframe classification method. The ultrasonic video classification work based on three-dimensional convolution requires the same number of video frames for different patients, and the efficiency of extracting features and the classification performance of the model are poor. In this work, We designed the ultrasound video classification method, realizes the training of ultrasound datasets with different video frame numbers, which increases the utilization rate of datasets, optimizes Spatio-temporal feature extraction, and improves classification accuracy.

\textbf{Methods:}
This paper proposed video classification methods based on CNN and LSTM. Moreover, to support variable-frame training, we introduce the processing scheme of long and short sentences in NLP into video classification for the first time. We reduce the dimension of the CNN extracted image features to 1x512 and then sort and compress the feature vectors and input them into LSTM for training. Specifically, the feature vectors are sorted by the number of patient video frames and populated with padding value 0 to form a variable batch. The invalid padding values in the batch are compressed and then entered into LSTM for training, saving computing resources. To calculate subsequent indicators, it is necessary to restore the output dimension to the input dimension by padding filling. 

To summarize, the contribution of our work is twofold: (a) we designed a set of CNN and LSTM training methods applied in the ultrasound video classification task. First, the CNN model is trained thoroughly, and the spatial feature vector of the ultrasound video is extracted. Then the LSTM is thoroughly trained with the feature vector as the input, and finally, the output probability of the LSTM is averaged to obtain the classification result. (b) To improve the utilization rate of the dataset, we introduced the solution of handling long and short sentences in NLP to the field of variable-frame video classification for the first time, which realizes the training of ultrasound datasets with different video frame numbers by sorting and compressing feature vectors, optimizing the spatiotemporal feature extraction, and improving the classification accuracy.

\textbf{Results:}
The equal frame and variable-frame CNNLSTM methods are superior to other methods in all metrics. Compared to the key frame classification method, the results show that the proposed method is better than the key frame method in terms of accuracy and precision. The specificity and F1 score both increased from 3\% to 6\%, and the specificity increased by 1.5\%. The accuracy rate, accuracy rate, and specificity of variable-frame CNNLSTM are improved compared to the equal-frame CNNLSTM. The above results confirm the proposed method's effectiveness.

\textbf{Conclusion:}
Experimental results demonstrate the proposed method's superior performance in classifying variable-frame ultrasound video. The developed variable-frame CNNLSTM can be further extended to the ultrasound video classification problems of other modalities of medical images.

\end{abstract}



\end{frontmatter}

\section{Introduction}
\label{sec:sample1}
Breast cancer is one of the most common malignant tumors in modern women, and its mortality rate has always been at the forefront of all malignant tumors worldwide. According to an epidemiological survey conducted by the international center for cancer prevention and research, changes in living habits, cultural background, and living environments, such as women delaying pregnancy and childbirth, sudden weight gain or long-term lack of regular physical activity and exercise, are the leading causes of breast cancer in women. The main risk factor for the sharp increase in cancer incidence. Early stage breast cancer has the potential to be cured~\cite{robertson2016essential}. Many scientific papers and clinically collected data have confirmed that examination and early diagnosis and treatment of diseases are the most critical methods to prolong patients' survival effectively. Therefore, early detection, early diagnosis, and early treatment are critical. This is a crucial premise for reducing the mortality rate of breast cancer~\cite{lee2002screening}, which can effectively improve the survival rate of female patients~\cite{chapman2007autoantibodies}. The most common imaging diagnostic methods in clinical practice include ultrasound imaging, CT imaging, and magnetic resonance imaging (MRI). Ultrasound imaging is widely used in most parts of the world due to its relative safety, low cost, non-invasiveness, and non-ionizing properties. It can assist radiologists in disease diagnosis. Many high-quality clinical images have been accumulated using advanced ultrasound technology, enabling large-scale deep model training.

The number of video frames of dynamic ultrasound data collected clinically is generally different, and existing ultrasound video classification methods require equal-frame datasets. For less-frame datasets, employing zero padding keeps the video frame length equal. The disadvantage of this processing is that when the filling value 0 is input into the LSTM for forwarding calculation, it wastes computing resources and may cause errors in the training results. For multi-frame data, employing the frame dropping method keeps frames equal, which leads to the waste of video frames. 

Current deep learning-based research focuses on ultrasound classification using single-frame ultrasound images. Since the timing characteristics of dynamic ultrasound video have a more significant impact on the diagnosis results in the clinical diagnosis of tumors, the current classification of dynamic breast ultrasound video is relatively lacking. The classification performance index of the existing method is not high. 

Ultrasound video is a precious data resource. It is necessary to develop a dynamic ultrasound video classification method with support variable frame training and better spatiotemporal feature extraction to assist doctors in ultrasound diagnosis. The proposed method can effectively reduce the workload of imaging doctors and the subjectivity of clinical diagnosis, thus making the diagnosis more accurate. This method consists of a convolutional neural network(CNN) and long short-term memory neural network(LSTM). First, the spatial features of the video frame are obtained through CNN, then the time-series features are extracted by combining with LSTM. The spatiotemporal features are combined to make the final classification prediction. Specially optimize the LSTM input structure, considering different video frame numbers to achieve batch training. This is conducive to extracting the time-series features of ultrasound videos of different lengths so clinical datasets can be fully characterized and applied. Combining the above innovative methods enables the entire model to obtain higher classification accuracy on dynamic ultrasound datasets. To summarize, the contribution of our work is threefold:  
\begin{itemize}
    \item We designed a set of CNNLSTM training framework for the ultrasound video classification task. First, the CNN model is trained thoroughly and extracts the spatial feature vector of the ultrasound video. Then the LSTM is thoroughly trained with the feature vector as the input. The output probability of the LSTM is averaged to obtain the classification result.
    \item The different methods used by clinicians to acquire ultrasound images result in different ultrasound video lengths. Therefore, we improved the CNNLSTM framework to propose the variable-frame video training method, which realizes the training of datasets with different video frame numbers by sorting and compressing feature vectors, optimizing the spatiotemporal feature extraction and improving the classification accuracy.
    \item Experimental results demonstrate the superior performance of the proposed method on ultrasound videos collected clinically. What's more, the dataset processing requirements of the proposed method are simple. No need to delineate the lesion area, just know the benign and malignant of the video frame. We will release part of the data to facilitate the research of computer-aided diagnosis of ultrasound images.
\end{itemize} 

\section{Related Work}
\label{sec:Related Work}
The classification task based on an ultrasound image is relatively extensive. The ultrasound classification task is mainly concentrated in the following fields: 1) Research on ultrasound image classification algorithms; 2) Research on ultrasound video algorithms. This section provides an overview of the relevant work in these two areas, and the limitations of the current method are described. 

\subsection{Ultrasound image classification}
With the development of deep learning technology, many researchers will use deep learning classifiers for ultrasound images to research ultrasound classification tasks. Singh~\emph{et al}.~\cite{singh2015adaptive} proposed a classification method for ultrasound images. First, we use a wavelet-based filter to remove blobs, and then the texture and shape features are extracted. Finally, use adaptive gradient descent for classification. Mohammed ~\emph{et al}.~\cite{mohammed2018neural} used a neural network approach for classification and used median and adaptive weighted filtering to preprocess images. Then, ROI and multifractal dimension features are extracted. Finally, the images were classified using an artificial neural network. Byra~\emph{et al}.~\cite{byra2018discriminant} used the extracted features of the VGG19 neural network architecture. Using Fisher discriminant analysis selects and classifies features. Fisher discriminant analysis identifies breast lesions and shows which features are helpful for contour detection. Yap~\emph{et al}.~\cite{yap2020breast} used a convolutional neural network (CNN) and a box classifier for line object recognition. First, convolutional features are extracted from ultrasound images and then objective using bounding boxes and object classification score. Finally, the box classifier identifies the tumor. Moon~\emph{et al}.~\cite{moon2020computer} proposed a combination of three different CNN classification structures. First, the original ultrasound images, ROI ultrasound images, tumor ultrasound images, tumor shape ultrasound images, and fused ultrasound images are extracted. Then, three CNN architectures (VGG, ResNet~\cite{he2016deep}, and DenseNet) are built from scratch and use machine learning algorithms to extract features. Next, combine the ensemble model with the CNN architecture. Finally, use an integration framework to classify. Huang~\emph{et al}.~\cite{huang2020segmentation} extracted the gray histogram, GLCM, and Symbiotic Local Binary Patterns (LBP). Using K-means and bag-of-words algorithms to extract features from GLCM and LBP. Then, an initial classification is performed using a Backpropagation Neural Network Work (BPNN) and K-Nearest Neighbors (KNN) algorithm for redistribution and post-processing. BPNN incorporates features from superpixels, while the KNN algorithm performs actual classification. Jarosik~\emph{et al}.~\cite{jarosik2020breast} proposed an ultrasound classification scheme based on radio frequency (RF) ultrasound signals. First, the RF patch is extracted from the original ultrasound image. The deep learning method then takes a 2D patch of the original RF signal and creates a sample. Next, concatenate the output vectors from the CNN global mean and max-pooling layers. Overall, the method involves three network architectures: the first applies global max-pooling to extract features, and the second consists of five blocks (2D convolution, max-pooling, average pooling, dense layers, and sigmoid activation), the third combines CNN-1D models with CNN-2D models. 
\subsection{Ultrasound video classification}
In clinical diagnosis, the time series in the ultrasound video has a positive effect on the diagnosis. Making good use of the time series in the video is the key to improving the accuracy of ultrasound classification. Bocchi~\emph{et al}.~\cite{bocchi2012semiautomated} proposed a breast ultrasound video classification algorithm that consists of five modules: preprocessing, semi-automatic segmentation, morphological feature extraction, and integration of each frame classification to obtain the final video classification result. The accuracy of the video algorithm is better than that of the single-frame ultrasound classification. However, the morphological features extracted from traditional methods only include shape, axial ratio, and echo features. Chen~\emph{et al}.~\cite{chen2021domain} proposed an ultrasound video classification method based on 3DCNN that combines clinical prior knowledge and ultrasound contrast technology. Because ultrasound contrast video can provide more detailed prior information, such as tumor blood supply, which extracts more comprehensive features. Compared to baseline methods~\cite{tran2015learning,tran2018closer,zhou2018temporal}, ~\cite{chen2021domain}can help classification models make more accurate diagnoses. However, contrast-enhanced ultrasonography is invasive detection, and an ultrasound contrast agent needs to be injected into the vein, and the price is also higher than B-mode ultrasound. B-mode ultrasound is safer and universal. In previous studies, time-series mining for B-mode ultrasound videos is insufficient. Therefore, this paper will explore and integrate ultrasound videos' spatial and temporal features based on the B-mode ultrasound videos collected clinically.

In order to further improve classification accuracy. We designed spatiotemporal feature fusion diagnostic methods with support variable frame input for the dynamic ultrasound video classification problem. This paper uses CNN and LSTM to extract spatiotemporal features from dynamic ultrasound data to improve classification accuracy. Specially optimize the LSTM input structure and achieve the different video frames batch training. The combination of innovative methods allows the entire model to obtain higher classification accuracy on the dynamic ultrasound dataset.

\section{Method}
\subsection{Overview}
\begin{figure}[ht]
\begin{center}
\includegraphics[width=0.90\textwidth]{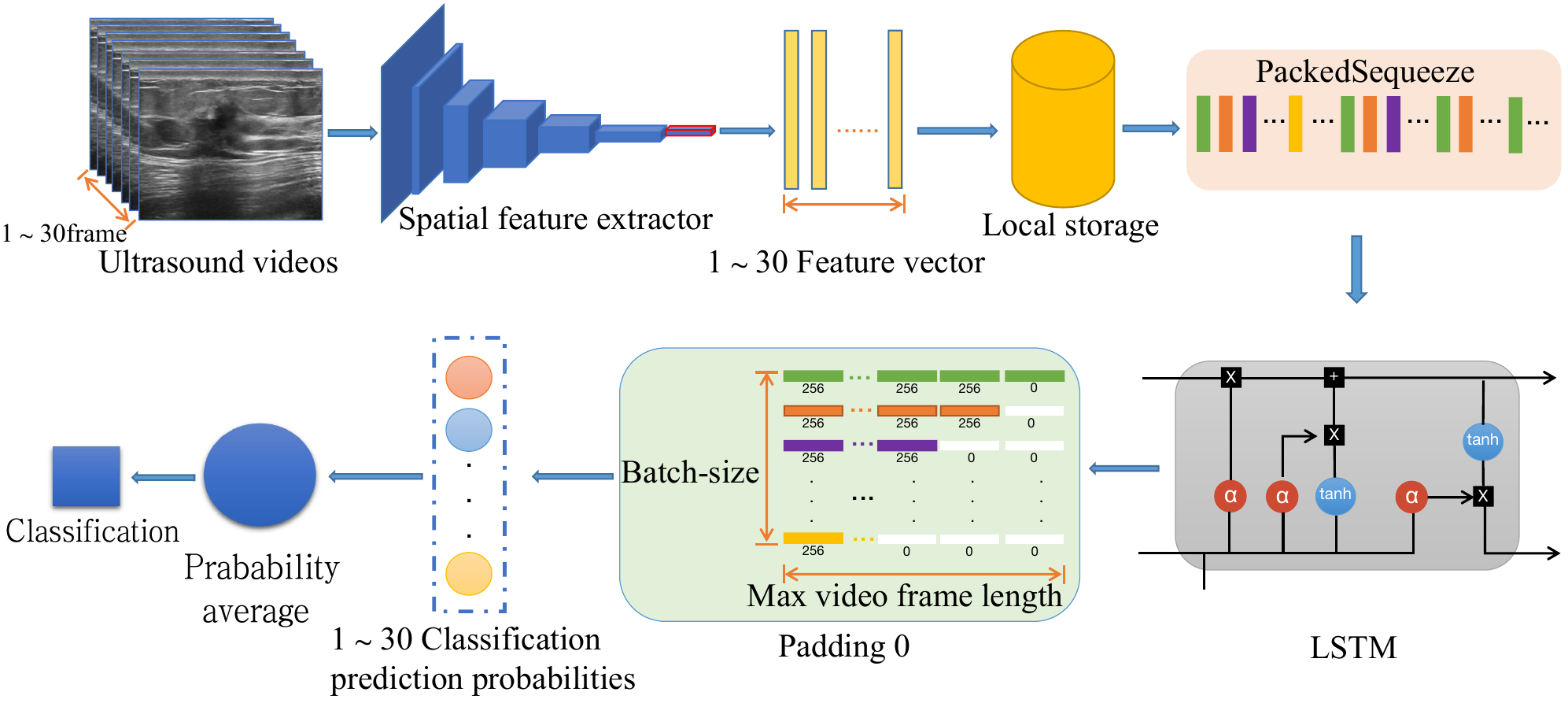}
\caption{Overview}
\label{fig: vs_overview}
\end{center}
\end{figure}

The overall structure of the ultrasound video classification method is shown in Fig.~\ref{fig: vs_overview}. The input is ultrasound video with 1 to 30 frames of different frame numbers. Using the spatial feature extractor to get 1 to 30 1×512-dimensional feature vectors. Then save the generated feature vector locally, and input it into LSTM after compression to get 1 to 30 1×256-dimensional prediction results. Finally, use softmax to get 1 to 30 classification prediction probabilities. The probabilities are added and averaged to get the final classification result.

\subsection{Spatial feature extraction}
\subsubsection{Loss function and training process}
With the rapid growth of ultrasound data, the ultrasound spatial feature vectors were extracted based on experience in the past. Generally, the model classification ability is weak when employing traditional methods to extract ultrasound morphological features, which is not enough to cope with the current large-scale data processing. In this paper, we use deep learning technology to realize ultrasound videos. To learn richer, more hierarchical, and higher quality ultrasound features to complete the ultrasound video classification work better. This section uses the deep residual network(ResNet) to extract ultrasound video features. In medical image classification tasks, the most commonly used loss function is the CrossEntropyLoss loss function. As shown in Eq.~\ref{eq: loss}:
\begin{equation}
\operatorname{loss}(\mathbf{x}, \text { label })=-\log \frac{e^{x_{\text {label }}}}{\sum_{j=1}^{N} e^{\mathbf{x}_{j}}}=-\mathbf{x}_{\text {label }}+\log \sum_{j=1}^{N} e^{\mathbf{x}_{j}}
\label{eq: loss}
\end{equation}

\begin{align}
H(X) &= H(X|Y) + H(X,Y) \\
&= -\sum_{x \in \mathcal{X},y \in \mathcal{Y}} p(x,y)\log p(x|y) - \sum_{x \in \mathcal{X},y \in \mathcal{Y}} p(x,y)\log p(x,y)
\end{align}

In Eq.~\ref{eq: loss}: x and label represent the sample and the true label, respectively.

The proposed method uses ResNet to extract spatial features of ultrasound video frames and then uses long short-term memory network LSTM to extract temporal features for classification. The specific process is: set the learning rate lr to 0.00001, the optimizer is Adam, and input 224 × 224 ultrasound images to the pre-training using ImageNet. The model will be trained for 300 rounds. In the Resnet model, the residual network outputs a feature vector with a length of 512. After three fully connected layers, the dimension of the feature vector is reduced to 2. To ensure the result's correctness, we use fully trained five-fold cross-validation. Finally, the model will extract the spatial feature vector of the dataset and save it locally as a npy file. 
\subsubsection{Spatial Feature Extraction Module}

\begin{figure}[ht]
\begin{center}
\includegraphics[width=0.90\textwidth]{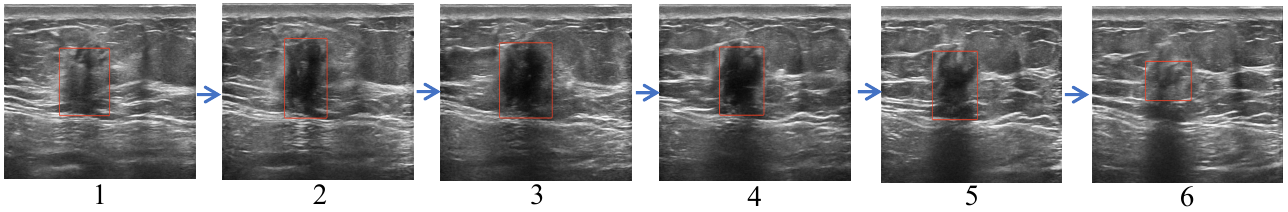}
\caption{Dynamic frame change}
\label{fig: vs_dynamic_frame_change}
\end{center}
\end{figure}

The time-series feature network training in Sec. 3.3 needs to use the spatial feature vectors extracted from the fully trained ultrasound classification model in Sec. 3.2.1 many times. Considering the saving computing resources, the spatial feature vector can be saved locally after being extracted and then used as the time-series feature extraction network's input for training to avoid wasting computing resources to extract the spatial feature vector multiple times. These saved feature vectors will be used as the input of the time-series feature extraction model LSTM to complete the model training. 

\begin{figure}[ht]
\begin{center}
\includegraphics[width=0.90\textwidth]{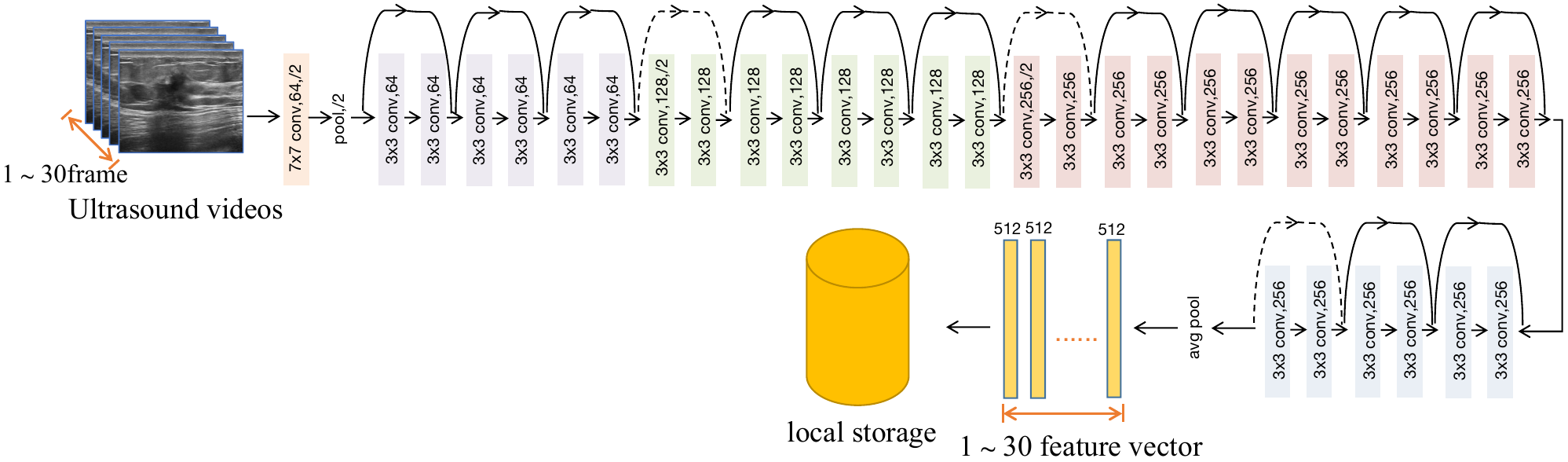}
\caption{Resnet framework}
\label{fig: vs_resnet_framework}
\end{center}
\end{figure}

As shown in Fig.~\ref{fig: vs_resnet_framework}, the specific method to extract the spatial feature vector of ultrasound video first loads the trained Resnet model weights and set the network to Eval mode. Then skip the ResNet fully connected layer in the forward to ensure that the final average pooling layer generates the extracted spatial feature vectors. 

\subsection{Time-series feature extraction}
In clinical diagnosis, as shown in Fig.~\ref{fig: vs_dynamic_frame_change}, doctors will analyze the dynamic changes of nodules through the ultrasound video stream, thereby improving the diagnosis accuracy.

In order to meet the clinical needs, we designed a time-series feature extraction method for ultrasound video. Fig.~\ref{fig: vs_overview} shows the whole process of the ultrasound video classification method. First, when extracting the depth features of ultrasound video, we fixed the video frame as a picture of size 224 × 224. The depth features extracted based on ResNet can reflect the spatial morphological features of ultrasound. Then, the feature vector extracted by ResNet is stored locally as a file. Then the feature vector will be input into the designed time-series feature extraction module to extract time series features further. Each video frame will have a separate prediction probability, and multiple video frames of a video will use the average prediction probability to determine the final video classification.

\begin{figure}[ht]
\begin{center}
\includegraphics[width=0.90\textwidth]{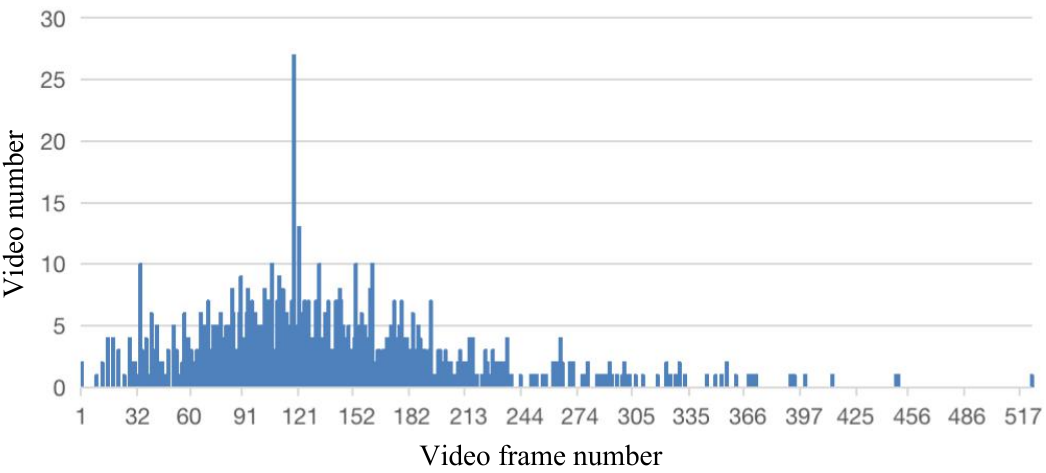}
\caption{Video frame number and video number}
\label{fig: vs_video_frame_number_and_video_number}
\end{center}
\end{figure}

\subsubsection{Time-series feature extraction module}
The current time-series feature extraction method is mainly extracted using 3D convolution, requiring the same video frame number in different patients and many parameters. As shown in Fig.~\ref{fig: vs_video_frame_number_and_video_number}, the abscissa is the frame number of ultrasound videos, and the ordinate is the number of videos corresponding to the number of video frames. In the field of medical images, because the time for clinically collecting ultrasound video is not easy to control, the number of video frames collected is often different. To use the data efficiently, the feature vector extracted by CNN is used as the input, and the time-series features are obtained by combining with LSTM. The proposed method improved classification accuracy and data utilization.

\subsubsection{Variable-LSTM module}
In the video classification task, LSTM is directly used to extract the video temporal features. However, the accuracy is low because LSTM cannot effectively extract spatial features. When the number of video frames is different, the method of zero-padding keeps the number of video frames the same. The disadvantage of this method is that employing the padding value 0 input into the LSTM for forwarding calculation, not only are computing resources wasted, but there may also be errors in results. 

\begin{figure}[ht]
\begin{center}
\includegraphics[width=0.90\textwidth]{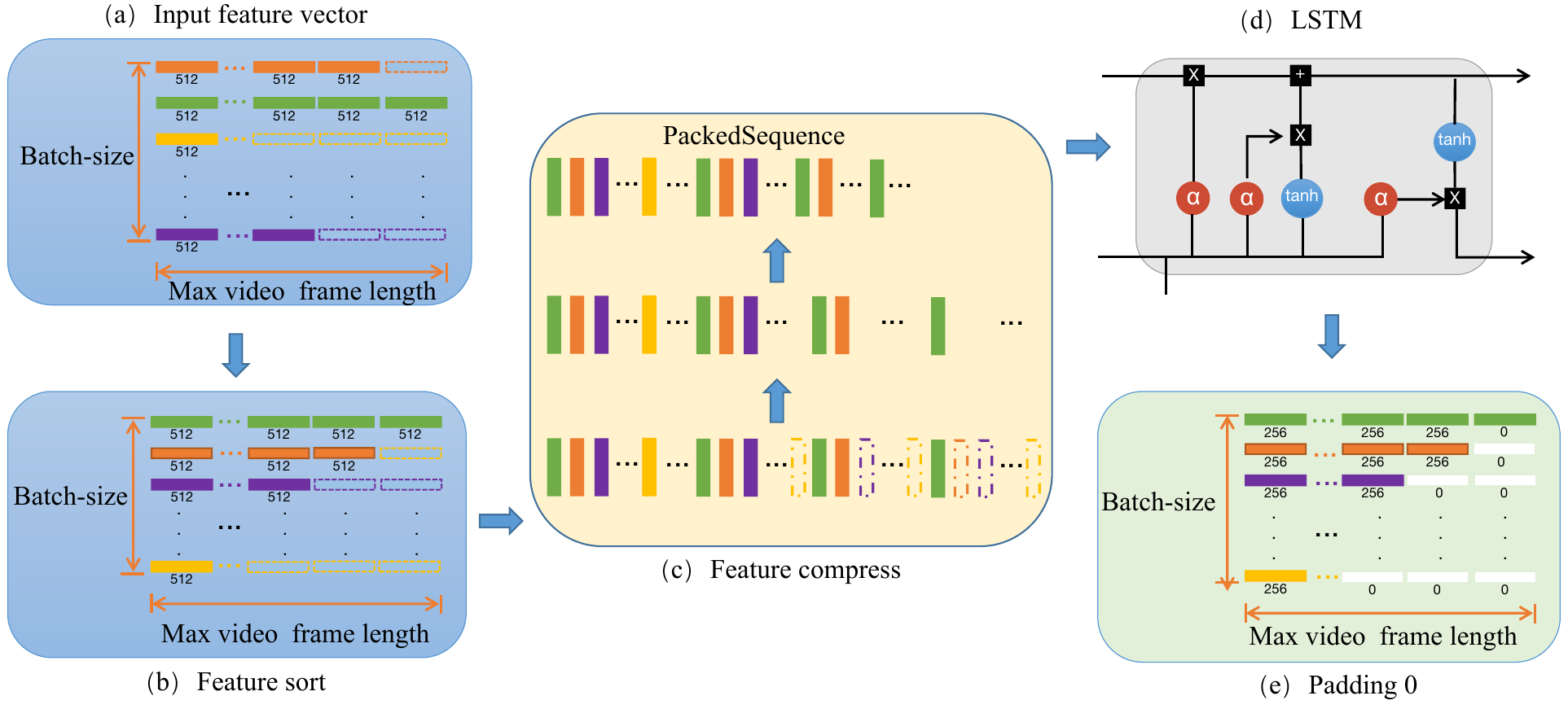}
\caption{Variable Lstm Framework}
\label{fig: variable_lstm_framework}
\end{center}
\end{figure}

Considering the above problems, in this chapter, the spatial feature vector extracted by CNN is passed as input to the variable-frame LSTM to extract time-series features. The classification is completed based on the extracted spatiotemporal feature vector. This proposed method can fully extract the spatiotemporal features of ultrasound videos with different number of video frames.

In natural language processing (NLP), the lengths of the sentences being processed are inconsistent. A standard method is to pad 0 after short sentences to keep the lengths of different sentences consistent. The disadvantage of this method is that padding 0 into LSTM for forwarding calculation wastes computing resources and may have errors in the final training results. The implementation principle of variable-frame LSTM shown in Fig.~\ref{fig: variable_lstm_framework} is mainly to expand the solution to similar problems in NLP. To perform batch training with variable-frames data more efficiently, the input data needs to be processed into PackedSequence format and then input to LSTM. It mainly involves three steps: vector sorting, compression, and zero-filling of prediction results to facilitate subsequent index calculation. 

As shown in Fig.~\ref{fig: variable_lstm_framework}, each row in (a) represents the feature vector of all video frames of a single patient, and each column represents the batch data. The feature vectors in (b) are sorted by the patient video frame number size. (c) shows the compression process of feature vectors. Specifically, the compression process is: the feature vectors in (b) are tiled into one dimension from left to right in units of columns. The PackedSequence format consists of compressed feature vectors and Batch-size. The predicted feature vector dimension output by the LSTM is 1×256. To facilitate the calculation of subsequent measurement indicators, zero-padding is performed on the prediction result (e), and the number of rows and columns of the prediction result (e) after zero-padding is consistent with the number of rows and columns in (b). 

\section{Experiments}
\subsection{Datasets}
Our data are breast ultrasound videos. They were collected and labelled by doctors in Xiangya Hospital. The label of the breast ultrasound image is whether the image contains nodules. As we all know, Xiangya Hospital is a very famous hospital. The breast ultrasound images collected and labelled by doctors in Xiangya Hospital have high reliability and are trustworthy. 
As shown in Table~\ref{tab: equal_frame_dataset_lists}, the ultrasound videos grouped by frame number have 10 frames, 12 frames, 14 frames, 16 frames, 18 frames, and 20 frames, which respectively contain 320 benign patients and 320 malignant patients. These different frame data are extracted from the original ultrasound videos using uniform sampling. 

\begin{table}[ht]
\caption{Equal video frame number dataset}
\begin{center}
\label{tab: equal_frame_dataset_lists}
{
\begin{tabular*}{\textwidth}{c@{\extracolsep{\fill}}clllll}
\hline
\multirow{2}{*}{Video Number} & \multicolumn{6}{c}{Frame Number} \\ \cline{2-7} 
                              & 10  & 12  & 14  & 16  & 18  & 20 \\ \hline
Benign                        & \multicolumn{6}{c}{320}          \\
Maligant                      & \multicolumn{6}{c}{320}          \\ \hline
\end{tabular*}
}
\end{center}
\end{table}

\begin{table}[ht]
\caption{Variable video frame number dataset}
\begin{center}
\label{tab: variable_frame_dataset_lists}
{
\begin{tabular*}{0.65\textwidth}{@{\extracolsep{\fill}}ccc}
\hline
\quad Frame Number & Benign & Malignant \quad \\ \hline
1$\sim$30    & 381    & 420       \\ \hline
\end{tabular*}
}
\end{center}
\end{table}

The variable frame ultrasound videos shown in Table~\ref{tab: variable_frame_dataset_lists} are also extracted from the original ultrasound videos using uniform sampling. The number of extracted frames is distributed between 1 and 30. The benign videos increased from 320 of the equal frame dataset to 381. The malignant videos increased from 320 of the equal frame dataset to 430. The dataset utilization rate has increased significantly.

\subsection{Datasets Preprocessing}
\begin{figure}[ht]
\begin{center}
\includegraphics[width=0.90\textwidth]{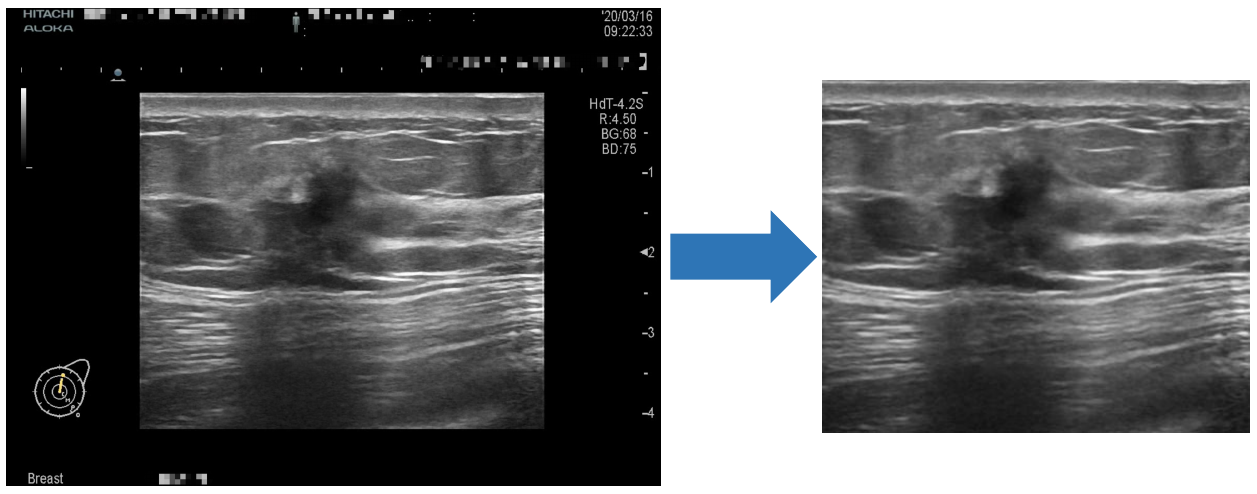}
\caption{Ultrasound image crop display}
\label{fig: vs_ultrasound_image_crop_display}
\end{center}
\end{figure}

In order to protect patient privacy and fully use clinically acquired ultrasound datasets, some necessary actions need to be taken to preprocess the datasets. Fig.~\ref{fig: vs_ultrasound_image_crop_display} shows the cropping operation. To protect privacy, the privacy information of the video frame on the left area has been hidden. There are many types and models of ultrasound data acquisition instruments, and there may be differences in hospitals and even departments. Therefore, after the private information is deleted through the data desensitization rules, the size of the ultrasound images may be inconsistent. To facilitate the deep model's training, the desensitized images' size is uniformly set to 224 x 224.

\subsection{Evaluation Metrics}
The confusion matrix plays an essential role in the performance evaluation of the classification model. The accuracy, sensitivity, and specific classification indicators such as sensitivity and specificity can be derived from the confusion matrix's TN, TP, FN, and FP. The TN stands for True Negative, which accurately shows the number of negative examples classified. Similarly, TP stands for True Positive, indicating the number of positive examples classified accurately. The term FP shows a false positive value, i.e., the number of actual negative examples classified as positive, and FN means a False Negative value which is the number of actual positive examples classified as negative. As shown in the following equations:

\begin{equation}
\text { Accuracy }=\frac{T P+T N}{T P+F P+T N+F N}
\label{eq: accuracy}
\end{equation}

\begin{equation}
\text { Precision }=\frac{T P}{T P+F P}
\label{eq: precision}
\end{equation}

\begin{equation}
\text { Sensitivity }=\frac{T P}{T P+T N}
\label{eq: sensitivity}
\end{equation}

\begin{equation}
\text { Specificity }=\frac{T N}{F P+T N}
\label{eq: specificity}
\end{equation}

\begin{equation}
F_{1} \text { Score }=\frac{2 \times \text { Sensitivity } \times \text { Precision }}{\text { Sensitivity }+\text { Precision }}
\label{eq: f1}
\end{equation}

In Eq.~\ref{eq: accuracy} to Eq.~\ref{eq: f1}: Accuracy represents the proportion of all correctly classified samples to the total number of samples. The precision indicates that the number of correctly classified positive samples accounts for the total number of positive samples proportion. The sensitivity indicates that the number of correctly classified positive samples accounts for the total number of correctly classified samples proportion. The specificity indicates the proportion of correctly classified negative samples to the total number of negative samples. The F1-score is calculated by sensitivity and precision, which is used to comprehensively judge the advantages and disadvantages of the model. 

\subsection{Experimental Settings}
For deep learning model training for dynamic ultrasound video classification, both CNN and LSTM models use the Adam optimization algorithm~\cite{kingma2013auto}, and the initial learning rate is $10^{-5}$. For the spatial feature extraction network CNN model, the video frame size is $224x224$. According to the GPU memory limit of the server, the batch size is 256, and the training cycle is 300 cycles. For the time-series feature extraction network LSTM model, the training period is 300, the input size is 512, the hidden size is 256, and the batch size is 32.

In order to fully train the CNN and LSTM network models, model performance is evaluated on a validation dataset every 20 epochs. The best performing model will be saved. In order to ensure the correctness, the experiment uses the five-fold cross-validation method. The five-fold method divides the dataset into five groups and conducts five independent training rounds. The specific process is:
\begin{itemize}
    \item Divide N pieces of data into several groups. Then take the $i^  {th}$ part as the $i^{th}$ validation dataset and the rest as the training dataset.
    \item Use DataLoader and Dataset to handle the training data in each group. Then take out data for training and testing in turn.
\end{itemize}
\subsection{Results \& Analysis}
\begin{figure}[ht]
  \begin{center}
  \includegraphics[width=0.95\textwidth]{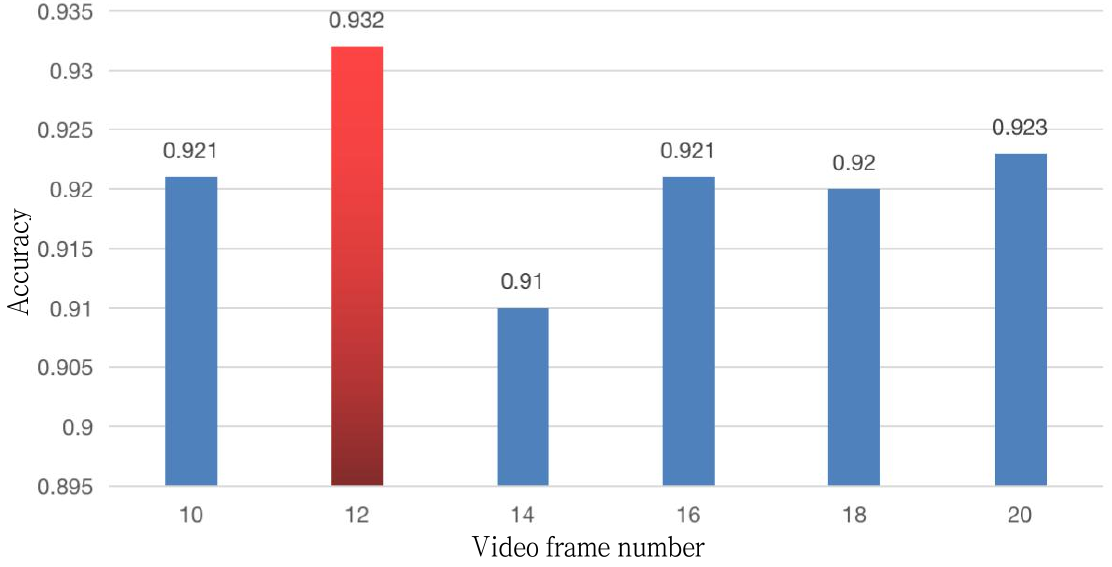}
  \caption{video frame parameter exp}
  \label{fig: video_frame_parameter_exp}
  \end{center}
\end{figure}

\textbf{parameter experiments:}
The proposed equal-frame CNN+LSTM and comparison methods require equal frame datasets, this section conducts parameter experiments on equal-frame datasets of 10, 12, 14, 16, 18, and 20 frames. The number of benign and malignant patient videos is 320 each. There are video frames lacking nodule features at both ends of datasets. Therefore, some frames will reduce the accuracy rate due to fewer nodule features. The experiment results in Fig.~\ref{fig: video_frame_parameter_exp} show that the 12 frames dataset has the highest accuracy rate. Therefore, all the comparison experiments utilized 12 frames dataset.

\textbf{Ablation Experiment:}
using 2DCNN extracted different video frame features. The classification probability is obtained by passing the feature vector generated by CNN into a fully connected layer and sigmoid. Next, there are two ways to complete the ultrasound video classification.
\begin{itemize}
    \item The classification of all video frames is statistically voted, and the classification type with the highest votes is the classification of the video.
    \item The classification probabilities of all video frames are added up and averaged, and then get the classification result.
\end{itemize}

As shown in Tab.~\ref{tab: ultra_vid_classification_ablation_exp}, Compared with CNN\_Vote and CNN\_Aver, the variable-frame CNN+LSTM method designed in this section improves accuracy, precision, specificity, and F1 indicators. It shows that the time-series feature vector extracted by LSTM supplements the spatial feature vector extracted by CNN, effectively improving the classification accuracy of ultrasound video.

\begin{table}[ht]
\caption{Ultrasound Video Classification Ablation Experiment}
\begin{center}
\label{tab: ultra_vid_classification_ablation_exp}
{
\begin{tabular}{cccccc}
\hline
Methods           & Accuracy & Precision & Sensitivity & Specificity & F1      \\ \hline
CNN\_Vote         & 92.34\%  & 92.76\%   & 91.88\%     & 92.81\%     & 92.29\% \\
CNN\_Aver         & 92.50\%  & 93.59\%   & 91.25\%     & 93.75\%     & 92.38\% \\
Variable CNN+LSTM & 93.46\%  & 95.53\%   & 90.26\%     & 96.28\%     & 92.77\% \\ \hline
\end{tabular}
}
\end{center}
\end{table}
 
\textbf{Comparative experiment:}
\begin{itemize}
    \item \textbf{Key\_Frame:} A key frame refers to a frame with more diagnostic information selected by the clinician in the ultrasound video. One key frame is extracted from the video, and then a residual neural network is used to extract the feature vector of the key frame. Finally, the classification result of the key frame is obtained through passing the feature vector into the fully connected layer and sigmoid. The classification results of key frames are more accurate than previous 3D convolution-based ultrasound video classification methods.
    \begin{figure}[ht]
    \begin{center}
    \includegraphics[width=0.90\textwidth]{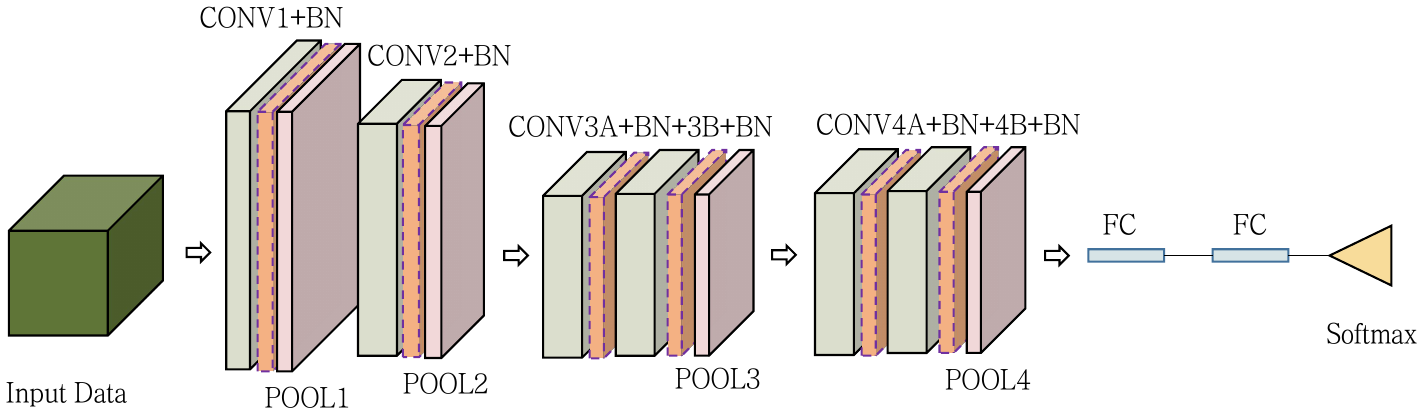}
    \caption{CNN C3D Framework}
    \label{fig: CNN_C3D_Framework_Display}
    \end{center}
    \end{figure}
    \item \textbf{C3D\_BN:} C3D~\cite{tran2015learning} designed a behavior recognition algorithm based on 3D convolution, which can effectively acquire and fuse multi-level spatiotemporal features from video data. Specifically, multiple information channels can be obtained from adjacent video frames, and then the information of all channels can be integrated to obtain the final classification result. In this section, C3D~\cite{tran2015learning} is applied to the ultrasound video classification task as a comparative experiment, but since the original C3D~\cite{tran2015learning} is based on 16 frames not 12 frames, the network structure needs to be modified. Some convolution, pooling layers and the input channels number of the fully connected layer are modified, etc. But the experimental results show that C3D~\cite{tran2015learning} performs poorly on the ultrasound video classification task. To further improve the classification performance of C3D~\cite{tran2015learning}, Batch\_Normalization, abbreviated as C3D\_BN, is added after each 3D convolution. As shown in Fig.~\ref{fig: CNN_C3D_Framework_Display}, the dashed box after each convolutional layer is Batch\_Normalization. Compared with the original C3D~\cite{tran2015learning} network graph, C3D\_BN reduces CONV5A+5B.
    \item \textbf{R3D:} To enhance the network expressive ability, the layer number of the network is further increased, but it may appear that the gradient disappears. The increased network layer does not perform better than the shallow network. To solve this problem, Resnet~\cite{he2016deep} introduced a deep residual framework to let the convolutional network learn residual mapping to solve this kind of network degradation problem. R3D~\cite{tran2018closer} replaces the convolutional network in C3D~\cite{tran2015learning} with Resnet~\cite{he2016deep}, which can effectively improve the performance of video classification. 
    \item \textbf{R(2+1)D:} 3D convolution can be approximated by 2D convolution and 1D convolution. R(2+1)D~\cite{tran2018closer} uses 2D convolution and 1D convolution to decompose space and time into two separate steps. The Relu activation function between 2D convolution and 1D convolution in each block makes the network nonlinear stronger than the 3D convolutional network using the same number of parameters. Therefore, the classification model can enhance the expressive ability of the network.
    
    \begin{figure}[ht]
    \begin{center}
    \includegraphics[width=0.90\textwidth]{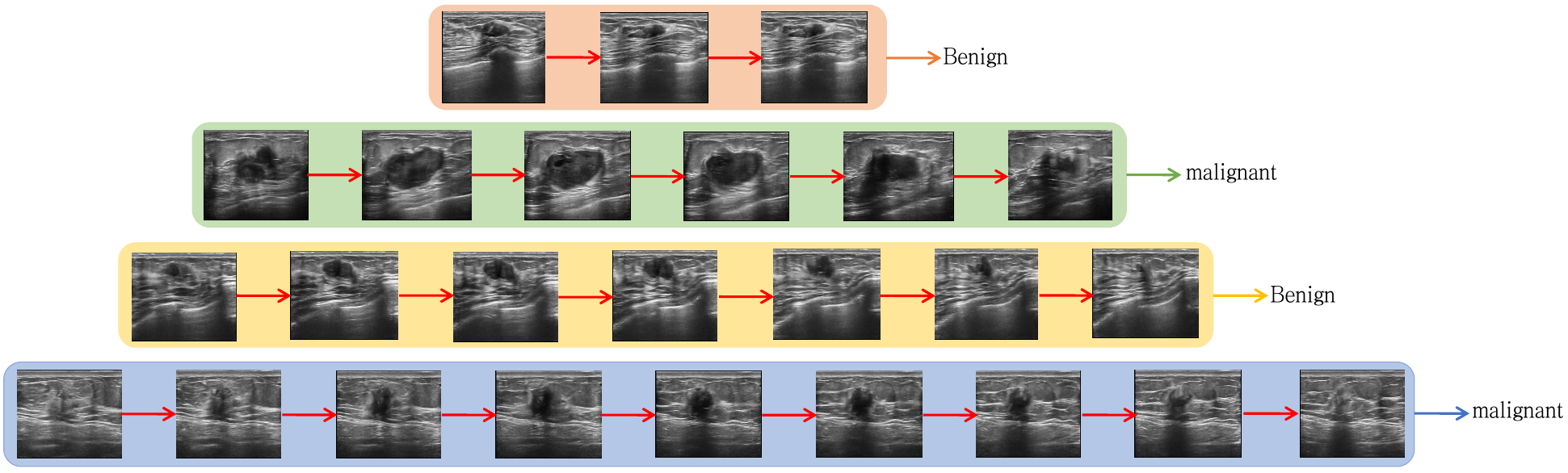}
    \caption{Virable Video Display}
    \label{fig: virable_video_display}
    \end{center}
    \end{figure}
    
    \item \textbf{Equal Frame CNN+LSTM:} This section combines CNN and LSTM to extract spatiotemporal features of video frames. First, fully train the CNN, then extract the spatial feature vector and save it to local. using feature vector as the input training dataset to train the LSTM and obtain the time-series features. Finally, obtain the classification by averaging the prediction probabilities of multiple video frames. Using uniform sampling to get equal frame data.
    \item \textbf{Variable frame CNN+LSTM:} The frame number of clinically collected ultrasound videos is often different. If satisfied equal frames requirements by zero padding, the classification accuracy will be affected and computing resources will be wasted. Therefore, we designed a CNN+LSTM ultrasound video classification method that supports variable frame training. Firstly, the spatial feature vectors extracted by CNN are sorted and compressed according to patients and then input into LSTM. The prediction results of LSTM will be padded with zero, which is convenient for subsequent metrics calculation.  
\end{itemize}

\begin{table}[ht]
\caption{Ultrasound Video Classification comparative Experiment}
\begin{center}
\label{tab: ultra_vid_classification_com_exp}
{
\begin{tabular}{cccccc}
\hline
Methods                & Accuracy & Precision & Sensitivity & Specificity & F1      \\ \hline
R(2+1)D~\cite{tran2018closer}                & 68.70\%  & 69.80\%   & 67.50\%     & 70.00\%     & 68.10\% \\
C3D\_BN~\cite{tran2015learning}                 & 70.46\%  & 69.17\%   & 76.88\%     & 64.06\%     & 71.94\% \\
R3D~\cite{tran2018closer}                    & 69.10\%  & 67.90\%   & 72.50\%     & 65.60\%     & 70.00\% \\
Key\_Frame~\cite{he2016deep}             & 89.22\%  & 88.95\%   & 89.69\%     & 88.75\%     & 89.26\% \\
Equal-frame CNNLSTM    & 93.13\%  & 94.84\%   & 91.25\%     & 95.00\%     & 92.97\% \\
Variable-frame CNNLSTM & 93.46\%  & 95.53\%   & 90.26\%     & 96.28\%     & 92.77\% \\ \hline
\end{tabular}
}
\end{center}
\end{table}

As shown in Tab.~\ref{tab: ultra_vid_classification_com_exp}, the equal frame and variable-frame CNN+LSTM methods are superior to C3D\_BN~\cite{tran2015learning}, R3D~\cite{tran2018closer}, and R(2+1)D~\cite{tran2018closer} in all metrics. The accuracy is increased from 22\% to 24\%, the precision is increased from 25\% to 27\%, the sensitivity is increased from 13\% to 23\%, the specificity is increased from 25\% to 32\%, and the F1 is increased from 20\% to 24\%. Compared to the key frame classification method, the results show that the proposed method is better than the key frame method in terms of accuracy and precision. The specificity and F1 score both increased from 3\% to 6\%, and the specificity increased by 1.5\%. The accuracy rate, accuracy rate, and specificity of variable frame CNN+LSTM are improved compared to the equal frame CNN+LSTM. The above results confirm the proposed method's effectiveness. The dynamic changes of nodules can be found in the fig.~\ref{fig: virable_video_display}. The variable frame video training and prediction can be achieved through the proposed variable frame CNN+LSTM method.

\subsection{Visual Analysis}
\begin{figure}[!ht]
  \begin{center}
  \includegraphics[width=0.70\textwidth]{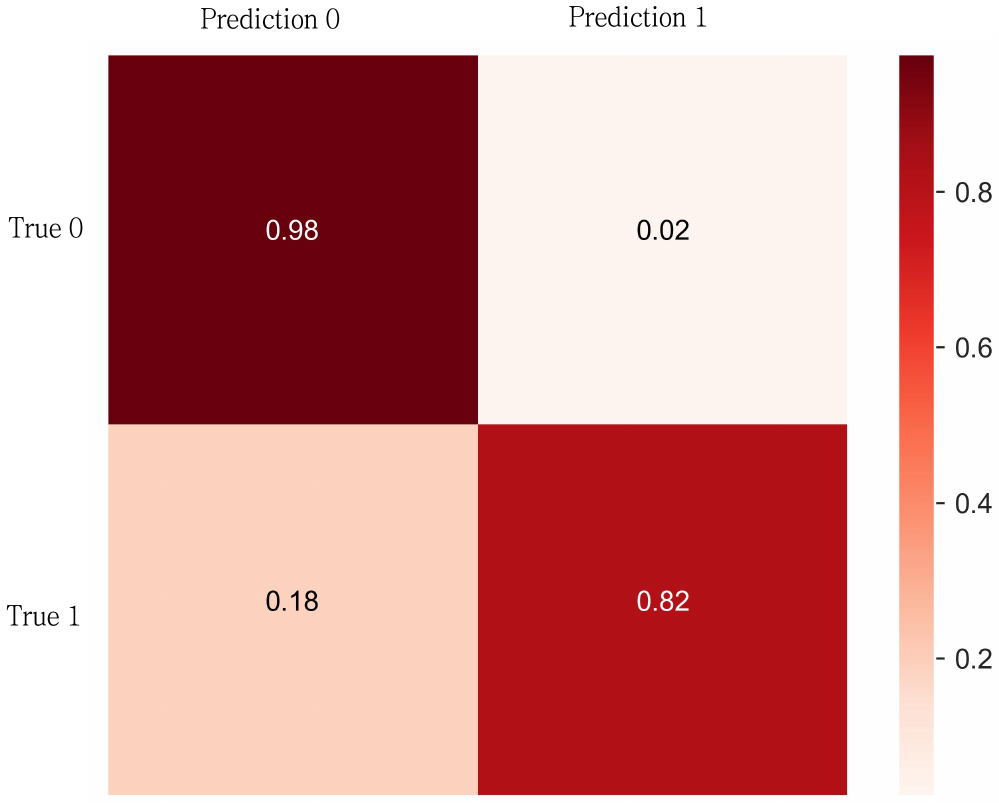}
  \caption{Normalized confusion metric}
  \label{fig: vs_normalized_confusion_metric}
  \end{center}
\end{figure}
\begin{figure}[!ht]
  \begin{center}
  \includegraphics[width=0.70\textwidth]{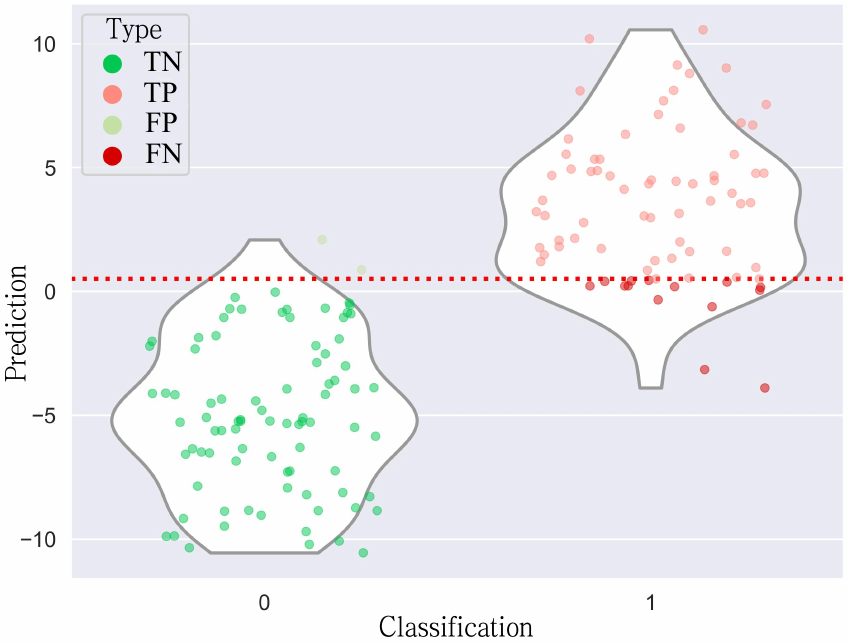}
  \caption{TP TN FN FP Distribution}
  \label{fig: vs_tp_tn_fn_fp_distribution}
  \end{center}
\end{figure}
\begin{figure}[!ht]
  \begin{center}
  \includegraphics[width=0.70\textwidth]{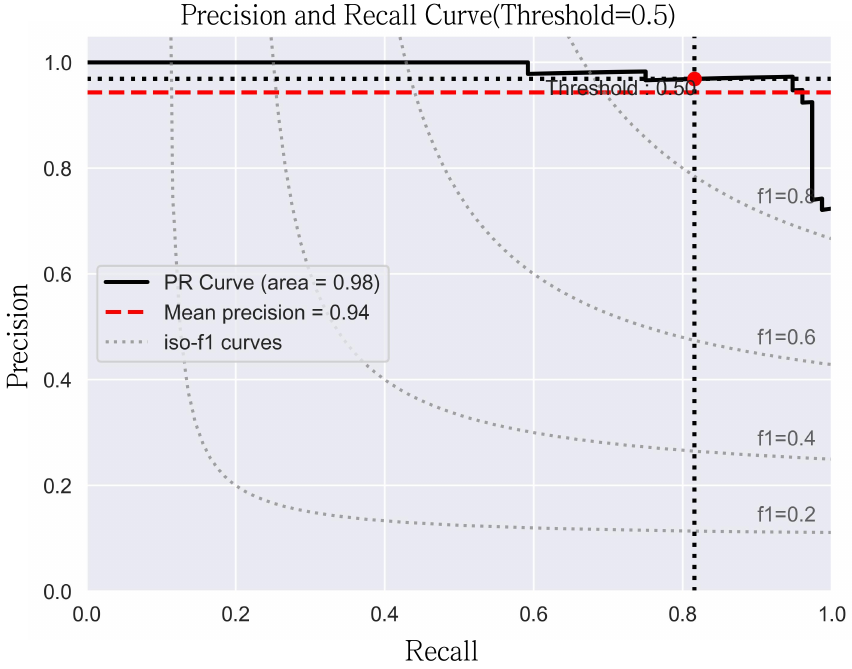}
  \caption{PR Curve}
  \label{fig: vs_pr_curve}
  \end{center}
\end{figure}
\begin{figure}[!ht]
  \begin{center}
  \includegraphics[width=0.70\textwidth]{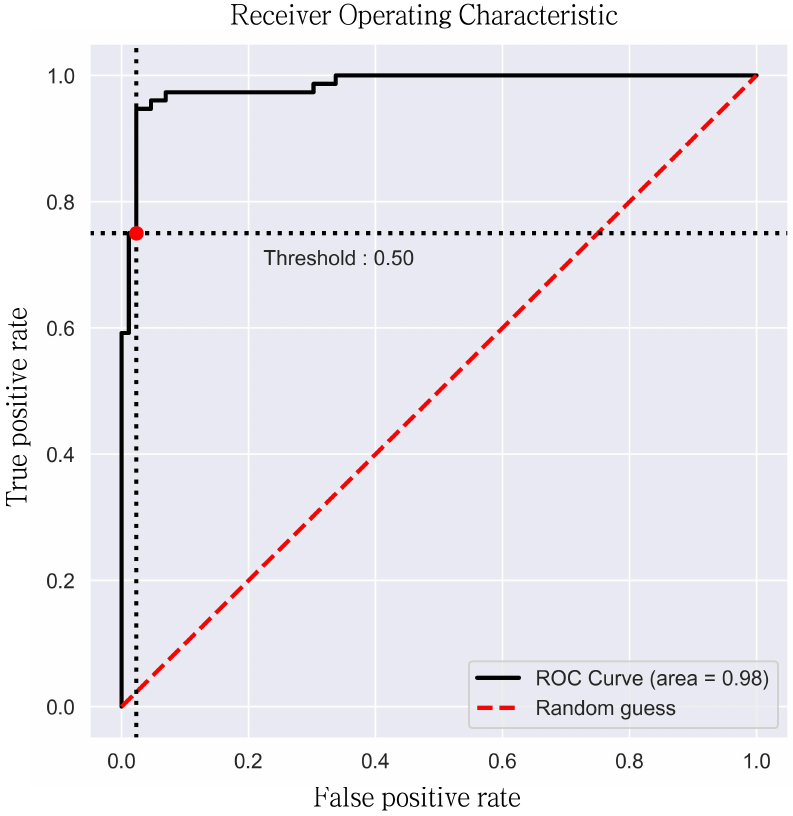}
  \caption{ROC Curve}
  \label{fig: vs_roc_curve}
  \end{center}
\end{figure}
In order to comprehensively analyze the performance of the variable frame CNNLSTM classification algorithm, this section obtains the confusion matrix result based on the threshold=0.5. As shown in Fig.~\ref{fig: vs_normalized_confusion_metric}, the output “TN” stands for True Negative, which accurately shows the number of negative examples classified. Similarly, “TP” stands for True Positive, which indicates the number of positive examples classified accurately. The term “FP” shows a false positive value, i.e., the number of actual negative examples classified as positive, and “FN” means a False Negative value which is the number of actual positive examples classified as negative. The TN is 82\%, the FN is 18\%, the TP is 98\%, and the FP is 2\%. However, fig.~\ref{fig: vs_normalized_confusion_metric} does not show the classification. To clearly illustrate the probability distribution of the classification, as shown in fig.~\ref{fig: vs_tp_tn_fn_fp_distribution} at Threshold=0.5 TP, TN, FP, FN distribution. fig.~\ref{fig: vs_tp_tn_fn_fp_distribution} can be found that the number of FN errors is more than FP and FN error scores. Most classes are near the boundaries of the two classes, so it is possible to classify these classifications. For example, through data enhancement to correct classification.

The PR curve plots precision versus recall at different thresholds, where the vertical axis denotes the precision score and the horizontal axis corresponds to the recall score. For the ultrasound video classification model, a specific point on the PR curve represents particular precision and recall values. The result is positive if the prediction classification probability exceeds the threshold. Otherwise, the result is negative. Then form a confusion matrix and calculate this specific point's recall value and precision value. However, only the precision and recall value measured at a certain point is not enough to comprehensively evaluate the model performance. It is necessary to consider the precision and recall values under multiple probability thresholds to analyze the entire PR curve trend. It can make an objective and comprehensive evaluation of the model. The entire PR curve is formed by taking multiple probability thresholds from [0, 1], then calculating the precision and recall scores under different probability thresholds and plotting them on the two-dimensional coordinate axis. The more probability thresholds are sampled, the smoother the PR curve. As shown in Fig.~\ref{fig: vs_pr_curve}, when the recall score is in the range of [0, 0.6], the precision is 1. When the recall rate is in the range of [0.6, 0.9], the area under the PR curve is 0.98, and the average precision is 0.94, which fully shows that the ultrasound video classification model of variable-frame CNNLSTM is effective. 

The receiver operating characteristic curve(ROC curve) is based on the true positive rate as the ordinate and the false positive rate as the abscissa. The ROC curve was used to evaluate the performance of the classification model at different classification thresholds. If the area is 0.5, the model is randomly classified and performs poorly. The classification model performance is powerful if the area is close to 1. As shown in fig.~\ref{fig: vs_roc_curve}, when the false positive rate is between 0 and 0.4, the true positive rate is more significant than 0.95. When the false positive rate is greater than 0.4, the true positive rate is 1. The area under the line of the ROC curve is 0.98, indicating that the classification performance of variable-frame CNN+LSTM is superior.

\section{Conclusion}
In the past, most ultrasound diagnostic algorithms were based on single-frame image diagnosis, ignoring the timing features, resulting in low classification accuracy. Moreover, the 3D convolution-based ultrasound video classification method that can extract spatiotemporal features requires the same number of video frames for different patients. However, the efficiency of feature extraction and the model classification performance is not good. In this paper, we proposed an ultrasound video classification method named variable-frame CNNLSTM, which supports variable video frame batch training. In order to protect patient privacy, using cropping to data desensitization. Then, a uniform sampling method extracts an equal-frame and variable-frame dataset from the original video. A detailed comparison and ablation experiments confirm the superiority of variable-frame CNNLSTM. After experimental comparison, the proposed method accuracy is improved by 4.24\% compared to the advanced method. The proposed method also increased ultrasound data utilization. The equal-frame and variable-frame dataset extraction method is uniform sampling in the ultrasound video classification task. However, this sampling method has a flaw. Some sampled frames lack classification characteristics, which will affect the classification results. In the future, two improvements: consider detecting the key-frame with the object detection method and then extracting the equal and variable frame datasets centered on the key-frame position. Some sampled frames lack classification characteristics, which will affect the classification results.

\section*{Acknowledgement}
This work is supported by the National Natural Science Foundation of China under grant 61902310 and the Natural Science Basic Research Program of Shaanxi under grant 2020JQ-030, and the Key Research and Development Program of Shaanxi under Grant 2021GXLH-Z-097.

\bibliography{cas-refs}

\end{document}